\documentclass{article}

\usepackage{PRIMEarxiv}

\usepackage[utf8]{inputenc} 
\usepackage[T1]{fontenc}    
\usepackage{hyperref}       
\usepackage{url}            
\usepackage{booktabs}       
\usepackage{amsfonts}       
\usepackage{nicefrac}       
\usepackage{microtype}      
\usepackage{lipsum}
\usepackage{fancyhdr}       
\usepackage{graphicx}       
\graphicspath{{media/}}     

\title{Efficient Whole Slide Image Classification through Fisher Vector Representation
}

\author{
  Ravi Kant Gupta, Dadi Dharani, Shambhavi Shanker, Amit Sethi \\
  Department of Electrical Engineering \\
  Indian Institute of Technology Bombay \\
  Mumbai, India\\
  \texttt{\{ravigupta131, 20d070023, 21d070066, asethi\}@iitb.ac.in} \\
}

\begin{document}
\maketitle

\begin{abstract}
The advancement of digital pathology, particularly through computational analysis of whole slide images (WSI), is poised to significantly enhance diagnostic precision and efficiency. However, the large size and complexity of WSIs make it difficult to analyze and classify them using computers. This study introduces a novel method for WSI classification by automating the identification and examination of the most informative patches, thus eliminating the need to process the entire slide. Our method involves two-stages: firstly, it extracts only a few patches from the WSIs based on their pathological significance; and secondly, it employs Fisher vectors (FVs) for representing features extracted from these patches, which is known for its robustness in capturing fine-grained details. This approach not only accentuates key pathological features within the WSI representation but also significantly reduces computational overhead, thus making the process more efficient and scalable. We have rigorously evaluated the proposed method across multiple datasets to benchmark its performance against comprehensive WSI analysis and contemporary weakly-supervised learning methodologies. The empirical results indicate that our focused analysis of select patches, combined with Fisher vector representation, not only aligns with, but at times surpasses, the classification accuracy of standard practices. Moreover, this strategy notably diminishes computational load and resource expenditure, thereby establishing an efficient and precise framework for WSI analysis in the realm of digital pathology.
\end{abstract}


\section{Introduction}
In recent years, the field of digital pathology has experienced a significant transformation, predominantly driven by advancements in high-throughput whole slide imaging (WSI) technology. WSIs, essentially high-resolution digital scans of tissue slides, have become integral in pathology for diagnostic, educational, and research purposes. However, the comprehensive analysis of these large and complex images presents considerable computational challenges. The traditional approach of analyzing entire slides is not only resource-intensive but also inefficient, given that not all regions in a WSI are equally informative for diagnosis.

Recognizing these challenges, our research focuses on a novel approach to WSI classification. We propose a method that selectively identifies and analyzes only the most pertinent patches within the slides. This approach is grounded in the hypothesis that certain regions within a WSI hold more diagnostic value than others and that their targeted analysis can yield efficient and accurate classification results.

The cornerstone of our methodology is the application of fisher vector representation for feature extraction. Fisher vectors have been used in image processing and computer vision due to their effectiveness in capturing fine-grained details in image data while still offering considerable data compression compared to using all available samples~\cite{akbarnejad2021deep}. By applying FV representation to the selected patches, we aim to extract rich and discriminative features that are crucial for accurate pathology classification in a data and computationally-efficient manner.

Through this research, we aim to contribute to the field of digital pathology by providing a more efficient and scalable approach to WSI analysis. Our method not only has the potential to enhance diagnostic accuracy but also to significantly reduce the computational burden associated with the analysis of large-scale pathological data. To validate our claims we have used The Cancer Genome Atlas (TCGA) Lung Dataset~\cite{coudray2018classification,gupta2023egfr} and Camelyon17 dataset~\cite{bandi2018detection} which consist of whole slide images. Snapshots of sample images from both datasets are shown in Fig.~\ref{fig1}. The Camelyon17 dataset, essential for the Camelyon17 Challenge, has facilitated the development of automated methods for the detection of breast cancer metastases in lymph node WSIs. It includes 500 high-resolution, expert-annotated images from various centers, categorized into four classes (Negative, Isolated Tumor Cell (ITC), Macro-metastases, and Micro-metastases). The TCGA Lung dataset has been used to develop automated methods to detect epidermal growth factor receptor (EGFR)  mutations from H\&E stained WSIs. It comprises 159 slides divided into EGFR and non-EGFR classes.

Our stated goals were achieved by proposing an efficient feature representation of whole slide images to address the unique challenges of H\&E stained histology images. The performance evaluation was focused on accuracy, robustness, and generalization to surpass state-of-the-art techniques on the two benchmark datasets. Furthermore, the research explored potential cross-domain applications in medical image analysis and computer vision, offering promising advancements in practical unsupervised domain adaptation with the help of effective feature space representation of whole slide images with different Fisher vectors to achieve significant improvement.
\section{Related Work}
Learning or extracting robust and discriminative features is vital for image classification. These features can be of two types -- handcrafted features and features learned from the data itself. Typically, classification methods for whole slide images (WSI) are grouped based on various factors such as their pooling techniques, the assumptions underlying their models, or the specific challenges they aim to tackle~\cite{dimitriou2019deep,ibrahim2020artificial}. 

One of the main challenges in WSI processing is their size. A typical WSI may have more than 100,000 pixels in each direction. Because images of this large a size cannot be processed in one go by a convolutional neural network (CNN) or a vision transformer (ViT), the WSI is broken into patches (sub-images) of a more manageable size. In a multiple instance learning (MIL) framework, each WSI is typically conceptualized as a `bag'. In this context, an instance might be a randomly selected patch from the WSI with or without its corresponding (i, j) coordinates within the WSI. Variations of the standard MIL assumption have been effectively applied in WSI classification~\cite{hou2016patch,combalia2018monte,li2019refinenet,chen2019rectified}. 

Under the standard assumption in this context, it is inferred that each instance within the 'bag' carries a hidden label. A bag is classified as positive if, and only if, it contains at least one positive instance. The  advantage of this assumption lies in its ability to closely align patch-level labels with WSI-level labels. This assumption becomes particularly relevant in scenarios where labels can be directly applied to individual patches (such as in the case of a colon cancer phenotype) or when the label determination is reliant on localized information. However, this is not the case for all labels. To modify this assumption, taking inspiration from the embedding-based methods in multiple instance learning, a common practice is to extract a set of patches, such as $\{x_1, ..., x_n\}$, from a WSI. These patches are then processed through a convolutional neural network (CNN), denoted as f, resulting in a set of vectors $\{f(x_1), ..., f(x_n)\}$. These vectors are subsequently combined into a single vector, which essentially represents the encoded information of the WSI from which the patches $\{x_1, ..., x_n\}$ are derived. For the combination of these values, various pooling strategies have been suggested, including max pooling, global average pooling~\cite{zhou2016learning}, and a combination of k-min and k-max pooling ~\cite{durand2016weldon}. Research indicates that the selection of a pooling strategy significantly impacts performance, both for natural ~\cite{durand2017wildcat} and histopathological images ~\cite{couture2018multiple}. A method closely related to ours is found in the study of ~\cite{couture2018multiple} and ~\cite{akbarnejad2021deep}, where they ~\cite{couture2018multiple} process a large image by calculating various quantiles of $f(x)$, with $x$ being a randomly selected patch.  And ~\cite{akbarnejad2021deep} designed a special purpose dataloader to get the patches from whole slide images. For WSI encoding, ~\cite{huang2019convolutional} suggests using a generative model, such as a variational auto-encoder or BiGAN, trained on patches from WSIs. Each WSI is then divided into a patch grid, with each patch (or grid cell) fed into the encoder of the trained generative model. The resulting volumetric map effectively represents the WSI and can be input into a subsequent module, like a CNN, for final prediction.

To address the computational challenges of processing a massive number of patches, recent methods have introduced novel approaches to enhance MIL frameworks. For example, the Attention-Challenging Multiple Instance Learning (ACMIL)~\cite{zhang2024attentionchallengingmultipleinstancelearning} method introduces semantic and diversity regularization to improve the extraction of discriminative patterns from WSI patches, overcoming limitations in traditional attention-based MIL methods. Similarly, the Prompt-Guided Adaptive Model Transformation (PAMT)~\cite{lin2024promptguidedadaptivemodeltransformation} framework enhances MIL classification by adapting pre-trained models to the specific characteristics of histopathology data, narrowing the domain gap between natural images and WSIs through innovative prompt-guided transformations. Furthermore, the Pathology-Knowledge Enhanced Multi-instance Prompt Learning (PEMP)~\cite{qu2024pathologyknowledgeenhancedmultiinstanceprompt} approach utilizes both visual and textual prompts, leveraging task-specific visual knowledge to improve few-shot learning scenarios in WSI classification. This method aligns visual and textual features using a pre-trained vision-language model, addressing the challenge of limited training data and enhancing classification accuracy

The major limitation of MIL-based methods are their exhaustive processing of every patch (or cell) in the grid through the encoder network~\cite{lu2021data}. Considering a typical WSI contains hundreds of thousands of patches, this method incurs significant computational expense. Fisher vector (FV) representation for feature extraction
from these patches is known for robustness in capturing
fine grained image details~\cite{perronnin2010improving}. Fisher vectors are especially useful for combining information from multiple regions of a large image.
\begin{figure*} 
\centering
\includegraphics[height=3.75cm,width=13.5cm]{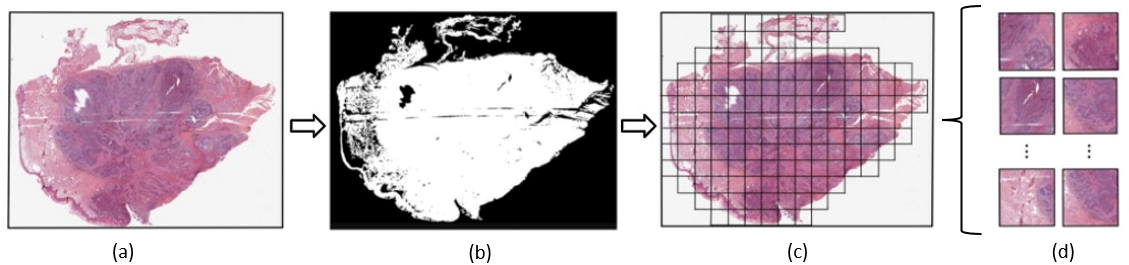}
\caption{Preprocessing framework: \textbf{(a)} represent thumbnail of WSI, \textbf{(b)} shows the tissue segmentation mask of WSI, \textbf{(c)} depicts patch generation from the tissue region of WSI, and \textbf{(d)} shows patches after filtering to get informative regions of WSI}
\label{prepro}
\end{figure*}

\vspace{1cm}

\begin{figure*} 
\centering
\includegraphics[height=4.5cm,width=15.5cm]{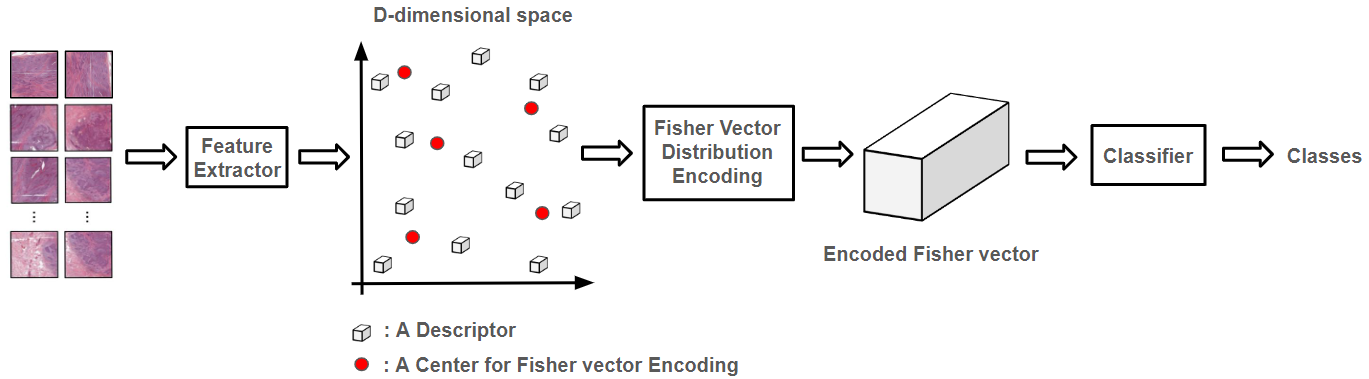}
\caption{Overall model framework}
\label{main}
\end{figure*}
\vspace{-10pt}
\vspace{-10pt}
\vspace{-10pt}
\section{Methodology}

WSIs are very large images whose each side can be of the order of 100000 pixels. Therefore, it is difficult to get representative patches from WSI which is capable enough to describe complete WSI. Our proposed method is mainly centered about to get a set of features as minimum as possible to represent the WSI, having the capability to describe the image in terms of classification task. To do so, we performed a set of preprocessing which includes, removal of artifacts using~\cite{PATIL2023100306}, tissue detection, patching, quality patches filtering, and nucleus count. To perform tissue detection we have used a python library HistomicsTK~\cite{HistomicsTK}  (shown in Fig.~\ref{prepro}(b)). After obtaining the tissue we performed patching using a sliding window algorithm (shown in Fig.~\ref{prepro}(c)). While applying the sliding window algorithm to get the patches, we extended this algorithm to filter out noisy patches. We extracted patches of size 512x512 pixels at 40x zoom level using OpenSlide library. For clean patches obtained from the previous process, we performed nucleus count using the library HistomicsTK for every patch to get the patches with high nuclei count. The core idea behind this is to get a more informative region of WSI (shown in Fig.~\ref{prepro}(d)). We are taking fixed number of patches ($x_1$, $x_2$, ........... $x_n$) of 512x512 pixel size from every whole slide image obtained after pre-processing as per compute availability. It will be more beneficial to have as many patches as possible. We are bound to have less number of patches because of computation bottleneck. This complete pipeline of preprocessing can be utilised as genuin practice to handle whole slide image. With the help of feature descriptor f(.) all patches are converted into D-dimenssion vectors as ($f(x_1), f(x_2), ........., f(x_n)$)~\cite{akbarnejad2021deep}.
Utilizing Fisher vector representation, we can intricately analyze the distribution of a set of all features, each existing in a d-dimensional space. This technique essentially extends the concept of a Gaussian mixture model (GMM) to encode the distribution of these high-dimensional features. In a Fisher vector framework, each feature is first associated with the parameters of the nearest Gaussian in the GMM~\cite{arun2020enhanced}. Subsequently, the gradients of the log-likelihood with respect to the parameters of these Gaussians are computed~\cite{arun2020enhanced}.

This process results in a Fisher vector for each feature, capturing the directional rate of change of the log-likelihood. By concatenating these vectors, we obtain a comprehensive and dense representation of the feature distribution. The dimensionality of the resulting Fisher vector is determined by the number of Gaussians in the mixture and the dimensionality of the features (D). Therefore, for a GMM with N Gaussians, the Fisher vector for a D-dimensional feature would have a dimensionality of 2ND (since it includes both mean and covariance information for each Gaussian)~\cite{novotny2015understanding}. The algorithm describes to get FV encoding below:

\noindent \emph{FV Feature Distribution Algorithm}

\noindent \textbf{Input:} Set of fixed d-dimensional features

\noindent \textbf{Output:} Comprehensive Feature Representation Vector

\begin{enumerate}
    \item Initialize Input Features (fixed number of features, each of d-dimensions).
    \item For each feature in the Input Features:
    \begin{enumerate}
        \item Associate the feature with the nearest Gaussian parameters in the Gaussian Mixture Model (GMM).
    \end{enumerate}
    \item Compute the gradients of the log-likelihood for each feature for the GMM parameters.
    \item Transform each feature's gradient into a Fisher vector.
    \item Concatenate all the Fisher vectors to form a single comprehensive feature representation vector.
    \item Output the comprehensive feature representation vector.
\end{enumerate}

In this method, we take into account a predefined set of vectors, denoted as $\{v_1,v_2....,v_M\}$, within the descriptor space. A collection of descriptors is  represented by $\{f(x_1), ..., f(x_n)\}$. The Fisher vector is generated by concatenating the normalized gradients of the log-likelihood of local image descriptors concerning the GMM parameters $\mu_{m}$ and $\sigma_{m}$~\cite{akbarnejad2021deep,arun2020enhanced}. The set of descriptors are encoded as:

\begin{equation}
FV(f(x_i)) = [\vartheta^{i}_{\mu,1},\vartheta^{i}_{\sigma,1},...,\vartheta^{i}_{\mu,M},\vartheta^{i}_{\sigma,M}]
\end{equation}
where FV: $\mathbb{R}^{(D+m)} \rightarrow \mathbb{R}^{2m}$,  and
\begin{equation}
\vartheta^{i}_{\mu,m} = \frac{\tau_m}{\sqrt{\pi_m}}(\frac{f(x_i)-v_m}{\sigma_m})
\end{equation}

\begin{equation}
\vartheta^{i}_{\sigma,m} = \frac{\tau_m}{\sqrt{2\pi_m}}(\frac{(f(x_i)-v_m)^2}{\sigma_m^2}-1)
\end{equation}
The $\mu_{m}$ and $\sigma_{m}$ denote the mean and standard deviation of the m-th Gaussian. The $\tau_{m}$ denotes the weight assigned to the descriptor $f(x_i)$, which is computed as follows:
\begin{equation}
\tau_m = \frac{\pi_m \mathcal{N}(f(x_i);\mu_m, \sigma_m^2)}{\sum_{j=1}^{M}\pi_j \mathcal{N}(f(x_i);\mu_j, \sigma_j^2)}
\end{equation}
where $\pi_{m}$ is the weight of the m-th Gaussian mixture such that $\sum_{m=1}^{M} \pi_{m}=1$. This method indicates the necessary adjustments to GMM parameters for a more accurate fit to the provided WSI descriptors.

The third segment of our model, average pooling, is a pivotal feature that ensures its general applicability across various WSI classification tasks. This technique effectively condenses complex data into a manageable form, maintaining the model's versatility and adaptability for diverse diagnostic scenarios in digital pathology. 

The final phase of our model is a classifier comprising multiple linear, fully connected layers that process the average pooled vector. This sequential arrangement of layers is meticulously designed to incrementally refine the input features, leading to the final class prediction. This structure highlights the effectiveness of linear transformations and deep neural networks in handling high-dimensional data, demonstrating their capability to extract and leverage complex features for precise classification in digital pathology. The whole pipeline of getting Fisher vector distribution and final classification result is described in Fig.~\ref{main}.
\\

\begin{table*}
\caption{ Results (\%) on the TCGA Lung dataset for binary classification.}
\begin{center}
\begin{tabular}{|l|c|c|c|c|c|c|c|}
\hline
 Method & Backbone Feature Extractor	 &	Accuracy &	 AUC	&	Precision & Recall & F1 score	\\
\hline
	DFVC~\cite{akbarnejad2021deep} & ResNet-50	 & 0.80	&	0.77	&	0.71&	0.68&	0.69\\
	
\hline
	Proposed Method & ResNet-50 &	\textbf{0.85} & 0.725	&	\textbf{0.75} &	0.6 &	0.67\\
\hline
Proposed Method & MobileNetV3small &	0.80 & \textbf{0.80}	&	0.67 &	\textbf{0.80} &	\textbf{0.73}\\
\hline
\end{tabular}
\end{center}  
\label{table1}
\end{table*}
\vspace{-15mm}
\begin{table*}
\caption{ Results (\%) on the Camelyon17 dataset for binary classification.}
\begin{center}
\begin{tabular}{|l|c|c|c|c|c|c|c|}
\hline
 Method & Backbone Feature Extractor	 &	Accuracy &	 AUC	&	Precision & Recall & F1 score	\\
\hline
	DFVC~\cite{akbarnejad2021deep} & ResNet-50	 & 0.75	&	0.73	&	0.71&	0.70&	0.70\\
	
\hline
	Proposed Method & ResNet-50  & \textbf{0.80}	&	0.72	&	\textbf{0.83}&	\textbf{0.85}&	\textbf{0.84}\\
\hline
Proposed Method & MobileNetV3small  & 0.72	&	\textbf{0.75}	&	0.80&	0.64&	0.71\\
\hline
\end{tabular}
\end{center}  
\label{table2}
\end{table*}
\vspace{10mm}
\section{DATASET AND IMPLEMENTATION DETAILS}
In our study, we utilized a modified ResNet50~\cite{jian2016deep} architecture and MobileNetV3small~\cite{howard2019searching} architecture trained on ImageNet~\cite{russakovsky2015imagenet} as our base model, where we removed the final pooling and fully connected layers. Subsequently, we integrated a 1x1 convolutional layer to compress the channel count, effectively reducing the descriptor space dimensionality to 10. This was followed by the addition of an instance normalization layer configured for 10 channels. For our experiments, we set the number of centers for Fisher vector coding (denoted as m in the equation) to five. Adhering to the notation in \cite{arun2020enhanced}, we configured the Fisher vector encoding parameters with $\pi_{m}$ at 0.2 and $\sigma_{m}$ at 0.1. An average pooling layer was then incorporated. The training process involved a batch size of 1 over 500 epochs, utilizing the AdamW optimizer with a learning rate of 0.00001 and weight decay of 0.00001.

\subsubsection{\textbf{Dataset}}
The Camelyon17 dataset, central to the Camelyon17 Challenge, is a pivotal resource in digital pathology, primarily focused on the automated detection of breast cancer metastases in lymph node WSIs. Comprising high-resolution images with expert-annotated metastatic regions, it serves as a standard for training and testing AI models in cancer diagnosis. The dataset presents a challenging task due to the variability in metastatic tissue appearance, making it a cornerstone for advancements in computational pathology.  This dataset, building upon its predecessor CAMELYON16~\cite{litjens20181399}, provides a diverse and realistic representation of clinical scenarios with detailed annotations by expert pathologists. These annotations include various metastasis categories, such as macro-metastases, micro-metastases, and isolated tumor cells, making them invaluable for developing and testing machine-learning models. This dataset consists of 500 WSIs from different centers having four classes (Negative, Isolated Tumor Cell (ITC), Macro-metastases, and Micro-metastases). We planned our experiments for two setups: one is a binary classification of Metastasis Positive vs Negative, and the other one is a four-class classification having classes Negative, ITC, Macro, and Micro. Additionally, our model underwent rigorous evaluation on the TCGA Lung dataset, specifically targeting variants in genetic mutations. Among these, the EGFR mutation stood out due to its aggressive nature, prompting a focused binary classification approach: EGFR mutations versus Non-EGFR mutations. This experimental design utilized a dataset comprising 159 slides, with 79 slides exhibiting EGFR mutations and the remaining 80 slides categorized under Non-EGFR mutations. This binary classification framework allows for precise discrimination between aggressive EGFR mutations and other genetic variants within the dataset.

\begin{figure*} 
\centering
\includegraphics[height=6.5cm,width=11cm]{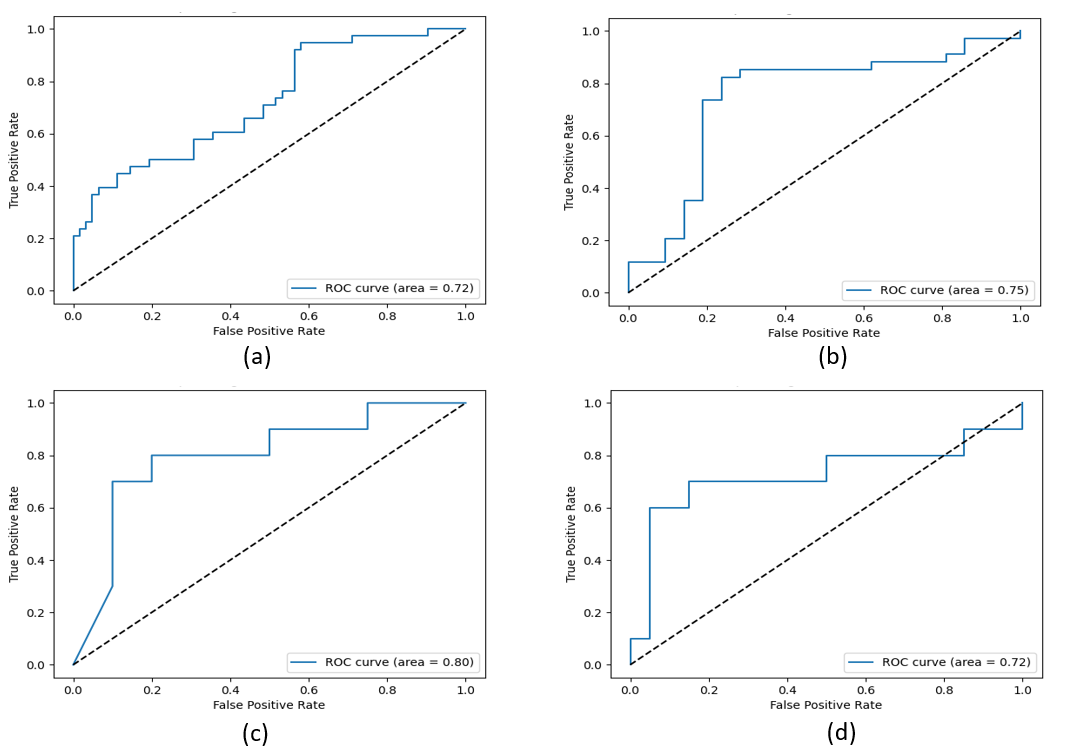}
\caption{Area Under the Curve (AUC) for binary classification for two datasets: Camelyon17 and TCGA Lung. Subfigures (a) and (c) show the results using the ResNet-50 model for the Camelyon17 and TCGA Lung datasets, respectively. Subfigures (b) and (d) depict the outcomes using the MobileNetV3small model for the same datasets}
\label{auc}
\end{figure*}
\section{Results}
Table~\ref{table1} illustrates the outcomes of the binary classification task on the TCGA Lung dataset, focusing on distinguishing between EGFR-positive and Non-EGFR samples, using the ResNet-50 model as the primary computational backbone. During the training phase, a consistent number of high-cellularity patches, selected post-filtering, were employed in each epoch. This approach yielded an accuracy of 0.85\% and an Area Under the Curve (AUC) of 0.73\%, with the corresponding AUC curve depicted in Fig.~\ref{auc} (c). This result surpassing the classical approach used in \cite{akbarnejad2021deep}. Our method also reduces the requirement of memory as we are prepossessing to get lesser patches to represent WSIs. An ablation study was also conducted for comparison, wherein a randomly chosen, fixed number of patches were utilized in each training epoch, to introduce variability in the whole slide image information. This method resulted in a lower accuracy of 0.67\%. 
When employing the same fixed number of patches in the Multiple Instance Learning (MIL) framework as ~\cite{gupta2023egfr}, the model exhibited signs of overfitting due to the limited dataset size, resulting in a notably poor accuracy of 0.52\%. This scenario underscores the challenge of model generalization within constrained data environments.
Additionally, the experimentation was extended using a smaller backbone model, MobileNetV3small, which surprisingly led to improved results with accuracy of 0.80\% and an AUC of 0.80\%, as shown in Fig.~\ref{auc} (d). This improvement, observed with the use of a larger number of fixed patches in combination with the smaller model, suggests that regions of high cellularity play a significant role in determining the EGFR class.

Table~\ref{table2} presents the classification outcomes on the Camelyon17 dataset, focusing on the binary differentiation of metastasis-positive versus metastasis-negative samples. In this analysis, the application of fixed high-cellularity patches in each training epoch, adhering to a specific filtering protocol, yielded an accuracy of 0.80\% and an Area Under the Curve (AUC) of 0.72\%, as depicted in Fig.~\ref{auc}(a). In contrast, an ablation study employing a random selection of fixed patches for each training epoch demonstrated a slightly higher accuracy of 0.81\%. These findings suggest that regions of high cellularity may not serve as reliable morphological biomarkers for metastasis detection. Furthermore, experiments utilizing the more compact MobileNetV3small model resulted in an accuracy of 0.74\% and an AUC of 0.75\%, as illustrated in Fig.~\ref{auc} (b).

In our study, we conducted a multi-class classification experiment using the Camelyon17 dataset, which includes 500 WSIs from five medical centers categorized into four pathological classes. We used five-fold cross-validation, training with slides from four centers and validating with slides from the remaining center in each fold. This method ensured robust and unbiased evaluation. The average classification outcomes were obtained using a model fine-tuned on the MobileNetV3small backbone. These results represent the composite average obtained from a cross-validation strategy where each fold involved using a different center (C0, C1, C2, C3) for training and the remaining center (C4) for testing, and then rotating such that C0, C1, C2, and C4 were used for training with C3 as the testing set in subsequent iterations. 

The method achieved an average accuracy of 0.68, peaking at 0.72 respectively, indicating the robustness of the MobileNetV3small model. In addition to the previously mentioned results, it is critical to note that centers C0 and C1 were indispensable during the training phase. Excluding these centers from the training set led to a significant imbalance in the data, adversely affecting the model's learning efficacy. The consistent inclusion of C0 and C1 in every training fold was crucial for maintaining a balanced distribution of classes across the dataset. This strategic choice ensured that the variance in classification performance across different training-testing splits was minimized, thereby highlighting the importance of a well-balanced training set for achieving reliable classification outcomes on the Camelyon17 dataset with the MobileNetV3small backbone.

\section{Conclusions and Discussions}
In our study, we introduced a new method that selects specific patches from
Whole Slide Images (WSIs) based on set criteria, unlike the conventional approach that uses all patches. This contrasts with Multiple Instance Learning
(MIL), which processes all available patches. Our method is end-to-end trainable, integrating patch selection and classification into a single process, unlike
MIL, where these steps are separate. We also used Fisher vector (FV) representation for WSI classification, comparing it with MIL. Our FV-based model is fully trainable for WSIs, unlike MIL, which treats feature extraction separately. A robust preprocessing pipeline, including data cleaning, feature engineering, handling imbalanced data, and domain-specific transformations, is crucial for our
model’s performance. These steps ensure cleaner and more relevant data, lead-
ing to more effective and reliable models. The FV method excelled in encoding
detailed information from WSIs, enhancing classification accuracy by capturing intricate image details. Its dimensionality reduction made processing more efficient, crucial for rapid clinical diagnoses. However, FV requires a pre-defined codebook and is computationally intensive, with performance sensitive to parameters and initial models, necessitating careful tuning. Conversely, the MIL approach treats each WSI as a bag of instances, focusing on the most discriminative patches. This method is effective for identifying rare but significant features
and handling unbalanced data, providing instance-level predictions. However,
MIL may overlook global contextual information, leading to suboptimal performance when the overall tissue structure is critical. Its effectiveness also depends on the quality and representativeness of selected instances. In summary, both FV and MIL have unique strengths and challenges in WSI classification. The choice depends on the task’s specific requirements, including dataset nature, desired detail level, computational resources, and clinical context. Future research could explore hybrid approaches, combining FV’s global context with MIL’s instance focus for more robust models in digital pathology. Our proposed method significantly reduces the computational burden by selectively filtering patches, allowing it to operate with fewer resources than the original state-of-the-art Fisher vector encoding method. Despite this resource efficiency, it achieves performance comparable to the current leading techniques. Given that our task is not traditional classification, an end-to-end pipeline is designed to perform the entire process seamlessly. Experimenting with different classifiers would disrupt this finely-tuned sequence and is not feasible within the current design framework, so we did not employ that kind of experimentation.

The future of Whole Slide Image (WSI) classification lies in hybrid models
that combine Fisher vector (FV) global analysis with Multiple Instance Learning (MIL) instance-level specificity. Using deep learning architectures like CNNs and Transformers can enhance feature extraction and spatial analysis. Techniques like Layer-wise Relevance Propagation (LRP) and Gradient-weighted Class Activation Mapping (Grad-CAM) will improve model transparency and interpretability. Our results highlight accuracy disparities across different centers, emphasizing the need for domain adaptation to ensure consistent performance. Future efforts will focus on developing methods to address these variations, setting new benchmarks in the generalizability and reliability of digital pathology models.

\bibliographystyle{unsrt}
\bibliography{references}

\end{document}